\documentclass[11pt,a4paper]{article}
\usepackage{times,latexsym}
\usepackage{graphicx}
\usepackage{booktabs}
\usepackage{tipa}
\usepackage{subcaption}
\usepackage{amsmath}
\usepackage{url}
\usepackage{tcolorbox}
\usepackage[T1]{fontenc}

\usepackage[acceptedWithA]{tacl2021v1}

\usepackage{xspace,mfirstuc,tabulary}

\newif\iftaclinstructions
\taclinstructionsfalse 
\iftaclinstructions
\renewcommand{\confidential}{}
\renewcommand{\anonsubtext}{(No author info supplied here, for consistency with
TACL-submission anonymization requirements)}
\newcommand{\instr}
\fi
\newcommand{\Table}[1]{Table~\ref{#1}}

\newcommand{\Appendix}[1]{Appendix~\ref{#1}}

\newcommand{\Figure}[1]{Fig.~\ref{#1}}

\newcommand{\Sectref}[1]{Sec~\ref{#1}}

\newcommand{\whisperipa}{Whisper}
\newcommand{\whisperipadist}{Whisper-PPT}
\newcommand{\fair}{XLS-R FAIR}
\newcommand{\ipapack}{\textsc{X-IPAPack}}
\newcommand{\nd}{XLS-R ND}
\newcommand{\dtest}{$\mathcal{D}_\text{test}$}

\usepackage{listings}
\usepackage{xcolor}

\usepackage{xcolor}

\definecolor{codegreen}{rgb}{0,0.6,0}
\definecolor{codegray}{rgb}{0.5,0.5,0.5}
\definecolor{codepurple}{rgb}{0.58,0,0.82}
\definecolor{backcolour}{rgb}{0.95,0.95,0.92}

\lstdefinestyle{mystyle}{
    backgroundcolor=\color{backcolour},   
    commentstyle=\color{codegreen},
    keywordstyle=\color{magenta},
    numberstyle=\tiny\color{codegray},
    stringstyle=\color{codepurple},
    basicstyle=\ttfamily\footnotesize,
    breakatwhitespace=false,         
    breaklines=true,                 
    captionpos=b,                    
    keepspaces=true,                 
    numbers=left,                    
    numbersep=5pt,                  
    showspaces=false,                
    showstringspaces=false,
    showtabs=false,                  
    tabsize=2
}

\iftaclpubformat 

\else

\fi

\title{Efficiently Identifying Low-Quality Language Subsets in Multilingual Datasets: A Case Study on a Large-Scale Multilingual Audio Dataset}

\author{
Farhan Samir$^{1,3}$\thanks{Work done while visiting the University of Washington.}~~~Emily P. Ahn$^{2}$~~~Shreya Prakash$^{2}$\AND Márton Soskuthy$^{1}$ ~~~ Vered Shwartz$^{1,3}$~~~Jian Zhu$^{1}$\\
$^1$ University of British Columbia\\
$^2$ University of Washington\\ 
$^3$ Vector Institute for AI\\
\texttt{fsamir@cs.ubc.ca}
}

\date{}

\begin{document}
\maketitle
\begin{abstract}

Curating datasets that span multiple languages is challenging. To make the collection more scalable, researchers often incorporate one or more imperfect classifiers in the process, like language identification models. These models, however, are prone to failure, resulting in some language subsets being unreliable for downstream tasks. We introduce a statistical test, the Preference Proportion Test, for identifying such unreliable subsets. By annotating only $20$ samples for a language subset, we're able to identify systematic transcription errors for $10$ language subsets in a recent large multilingual transcribed audio dataset, \ipapack{} \citep{zhu-etal-2024-taste}. We find that filtering this low-quality data out when training models for the downstream task of phonetic transcription brings substantial benefits, most notably a $25.7\%$ relative improvement on transcribing recordings in out-of-distribution languages. Our method lays a path forward for systematic and reliable multilingual dataset auditing.
\end{abstract}
\section{Introduction}
High-quality multilingual datasets are imperative for developing equitable language technologies. Access to large multilingual datasets is improving in some ways. Only in the last few years, researchers have open-sourced a number of notable open multilingual datasets, such as MADLAD-400 \citep{kudugunta2024madlad}, VoxPopuli \citep{wang-etal-2021-voxpopuli}, and OWSM \citep{Peng2024OWSMVB}.

The acquisition of these large multilingual datasets is complex, however. The data collection pipeline needs to scale to a large number of languages and a large volume of data for each language. Enabling this scaling is the use of predictive models: from language-identification models to sentence-embedding based bitext mining methods \citep{kreutzer-etal-2022-quality,koehn-etal-2020-findings} to speaker diarization models \citep{wang-etal-2021-voxpopuli}. 


These models however are prone to failure: Major data collection errors in well-known datasets have already been reported. In a comprehensive study over 5 major datasets and over $70$ languages, \citet{kreutzer-etal-2022-quality} report systematic failures in language identification and bitext mining, with greater error rates for lower-resourced languages. 
Further, recent studies in multilingual and long-form speech recognition found that prominent speech datasets contain substantial chunks of untranscribed content, leading to a high rate of deletion errors in models trained on those corpora \citep{tian2024effects,fox2024updated}.

\begin{figure}
    \centering
    \includegraphics[scale=0.30]{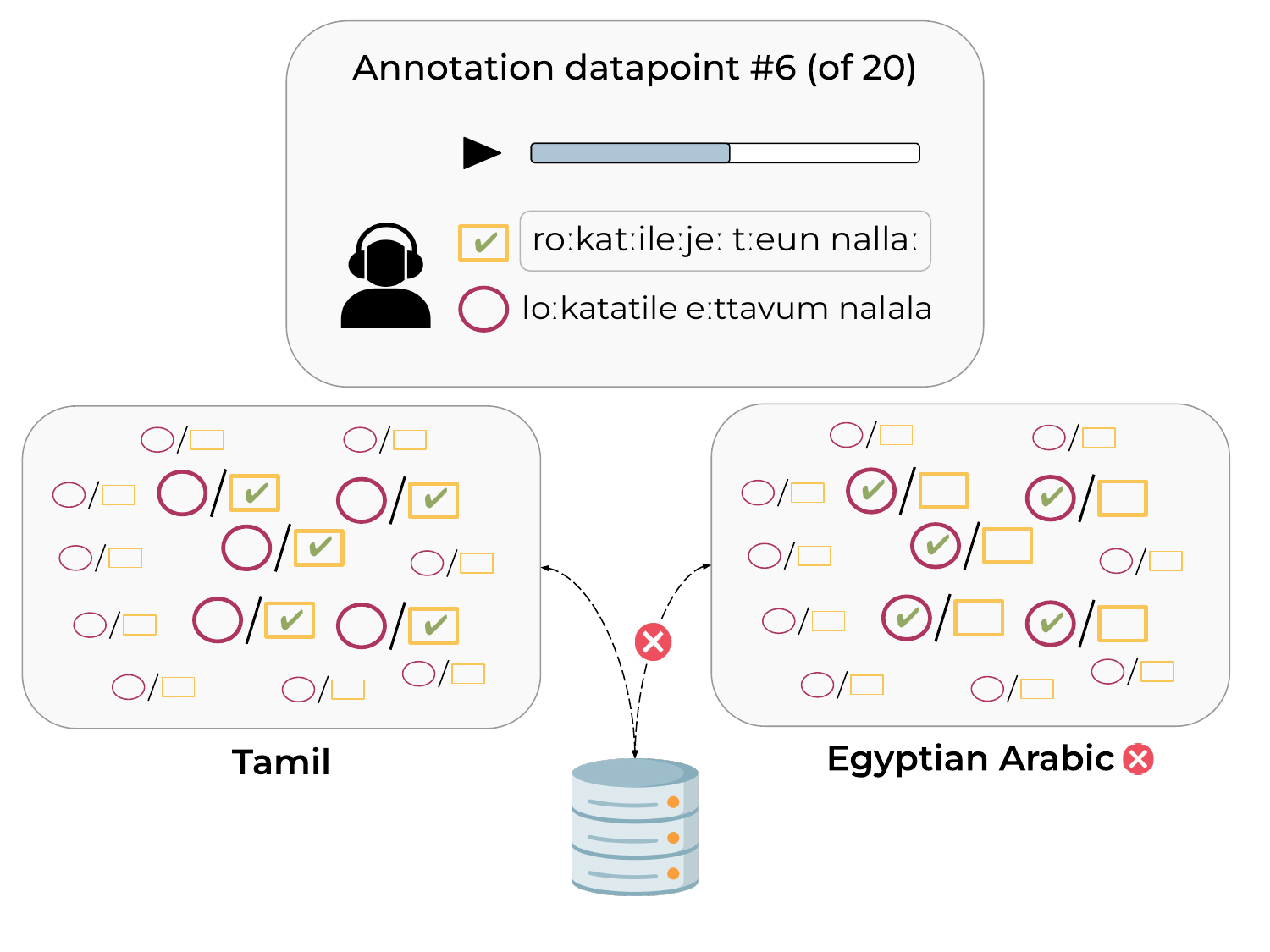}
    \caption{Multilingual datasets have been reported to contain a high degree of quality-control issues, especially for lower-resourced languages. We audit the quality of each language subset in a multilingual dataset by annotating a small sample from it. Specifically, expert annotators select between the \colorbox{yellow}{gold-standard transcript} and one generated by a \colorbox{magenta}{baseline model prediction}. When the baseline model predictions are consistently preferred over the gold-standard, we conclude that the language subset should be flagged for re-labeling.}
    \label{fig:preference}
\end{figure}

These studies demonstrate the complexity of acquiring high-quality multilingual data. In this light, the data collection pipeline itself can be considered an imperfect approximation of the data distribution we wish to sample from. As with any approximation, common wisdom suggests that we should aim to evaluate the quality of the approximation. Unlike the wealth of empirically and theoretically established metrics and hypothesis tests for comparing two models \citep{dror-etal-2018-hitchhikers}, there is a remarkable dearth of methods for evaluating the reliability of a semi-automatically scraped dataset that may serve as ``gold-standard'' for future downstream applications. 

In this work, we ask: How can we efficiently identify languages or dialects where the data-collection pipeline may have failed systematically? At a high level, we want to analyze a small subset of the data for a language from a multilingual dataset, and determine whether its samples are high quality. This analysis has two important components. First, how do we determine if a sample is high quality? Second, how large should the subset be to make our determination?

To answer these questions, we draw on theory from statistical power analysis \citep{cohen1992statistical,card-etal-2020-little}. Specifically, we are interested in accurately estimating \textit{preference} \citep{bradley1952rank}: how much better does the dataset capture the relationship of interest than an existing baseline model? In particular, we elicit preferences from human annotators to ground the dataset quality with judgments from domain experts.

We ground our method in the task of phonetic transcription, where the input is recorded speech from any language, while the output is a transcription into the International Phonetic Alphabet (IPA). Phone recognition models have important applications in language documentation, especially for oral languages \citep{bird2021sparse,lane2021local}. The training datasets for this task are often semi-automatically generated, so that many languages can be represented  (Sec.~\ref{sec:related-work}).

Consider the subset for one language in such a semi-automatically generated multilingual dataset. If a knowledgeable user consistently prefers the output of an existing (imperfect) baseline model over the ``ground-truth'' transcripts in the dataset under audit (see \Figure{fig:preference}), then this indicates that this language subset is unreliable. We refer to this statistical test as the Preference Proportion Test, or the PPT, which we introduce in \Sectref{sec:quality-audit}. Critically, we assert that only a small fraction of examples needs to be annotated to attain a high-powered test of whether a language subset should be flagged as unreliable.


To illustrate the effectiveness of the PPT, we perform a case study on a recent large multilingual phonetic transcript dataset -- the \ipapack{} dataset \citep{zhu-etal-2024-taste}, comprising transcribed audio for $78$ languages. Applying the PPT, we efficiently identify $10$ language subsets in the dataset that have unreliable transcripts. We find that a model finetuned on the filtered version of the dataset -- without the unreliable $10$ language subsets -- generalizes \textit{better} to a test set (comprising $5$ held out languages) than a model trained on the larger unfiltered dataset. 

We find two ways in which low-quality data can be especially pernicious. First, we find the largest improvement on the Punjabi subset of our held-out evaluation dataset  -- an error reduction of $20.3\%$  -- possibly due to omitting the unreliable transcripts from the Sindhi dataset, suggesting that the effects of poor data quality can tamper with performance in related languages.
Second, we also find a $25.4\%$ improvement on out-of-distribution languages after training on the PPT-filtered dataset, suggesting that low-quality data has a considerable impact on lower-resourced languages. Our empirical results add nuance to the purported benefits of data-scaling \citep[][for example]{hoffmann2022training}.


Finally, we emphasize that filtering out low-quality data, while highly effective, is not a panacea for building robust multilingual models. Some works have suggested that high-quality data on a small handful of languages is sufficient to obtain a ``universal'' or ``language-agnostic''  model \citep{taguchi2023universal,li2020universal}.Our empirical results in phonetic transcription however do not support this position. Leveraging a phone segment-level error metric, we find that existing universal phonetic transcription models are instead highly attuned to sounds that are more common in their respective training datasets, while making more errors on unfamiliar sounds. Overall, this suggests that more diverse and high-quality data collection is required for equitable performance across languages and their varieties. 

\section{Related work} \label{sec:related-work}

\noindent \textbf{Large multilingual speech corpora.} We chose to investigate the quality of multilingual phonetic transcript datasets in particular, as researchers have semi-automatically generated large datasets for well over $700$ languages. Among these, \citet{li2021multilingual} curated a multilingual speech corpus comprising audio data and transcriptions from $95$ languages, originally collected for the UCLA Phonetics Archive \citep{Ladefoged2011UCLAPL}. \citet{chodroff2024phonetic} further extended this work, providing fine-grained phone-level alignments of this corpus. \citet{ahn2022voxcommunis} employ Epitran \citep{mortensen2018epitran} to generate phonetic transcripts for the CommonVoice dataset. \citet{salesky2020corpus} generate phonetic transcripts for a large subset (of $635$ languages) from the CMU Wilderness corpus \citep{black2019cmu}. \citet{zhu-etal-2024-taste} generate phonetic transcripts using G2P tools for the Fleurs dataset \citep{conneau2023fleurs}. In this work we probe the \ipapack{} dataset introduced by \citet{zhu-etal-2024-taste}, as it is the most recent phonetic transcript dataset and it was used to train a high-quality keyword spotting model, suggesting that it has at least some high-quality data.

\noindent \textbf{Dataset auditing methods}. Validating the quality of ground-truth labels or transcripts in multilingual datasets is challenging, and is currently performed in ad-hoc ways, precluding community standards. \citet{kreutzer-etal-2022-quality}  perform their large-scale quality audit by creating a taxonomy of labels for making fine-grained categorizations within the broader dichotomy of correct/incorrect, and having annotators perform this classification. \citet{kudugunta2024madlad} inspect $20$ documents for each language in their dataset, excluding languages that contain ``majority noise''. However, they do not describe what constitutes noise. 

For assessing the quality of their phonetic transcript dataset, \citet{salesky2020corpus} produce average Mel-Cepstral Distortion scores for each language. They note that a large number of languages have a high distortion score; nonetheless, a high score is insufficient for determining whether the language's data is reliable, and thus they do not filter out any unreliable language partitions with this measurement.
For validating the quality of the \ipapack{} dataset, \citet{zhu-etal-2024-taste} assess $10$ samples per language and check that each sample is of reasonably high quality. However, as noted in work on Reinforcement Learning from Human Feedback \citep{Christiano2017DeepRL}, absolute judgments tend to be unreliable compared to relative comparisons of the form in \Figure{fig:preference}. Moreover, the choice of selecting only $10$ samples to validate per language is unsubstantiated. In \Sectref{sec:ipapack}, we revisit the quality auditing of the \ipapack{} dataset, improving upon both of the aforementioned limitations. Then in \Sectref{sec:experiments}, we assess the impact of our quality audit on the downstream task of automatic phone recognition.

\section{Case Study: Auditing \ipapack{}} \label{sec:ipapack}
In this section, we introduce the Preference Proportion Test for efficiently auditing the quality of a multilingual dataset. 
We ground our explanation in a case study of the \ipapack{} dataset, specifically the \ipapack{}-\textsc{Fleurs} partition.  
We first describe the \ipapack{} dataset and the preprocessing of the text in the phonetic transcripts (Sec.~\ref{sec:transcript-composition}).
Then, we introduce the test (Sec. ~\ref{sec:quality-audit}) and apply it to the \ipapack{} dataset (Sec. ~\ref{sec:ppt-application}).

\subsection{\ipapack{} Contents} \label{sec:transcript-composition}
\begin{table}[]
    \centering
    \begin{tabular}{p{1.2in}p{1em}p{3em}p{4em}}
         Category & Ex. & \#\{Type\} & \#\{Tokens\}\\
         \toprule
         Valid primary & p & $107$ & $21.8$M\\
         Valid one diacritic & $\text{v}^{\text{j}}$ & $282$ & $2.1$M \\
         Valid two diacritics & $\text{k}^{\text{hj}}$ & $67$ & $62.5$K\\
         Unlikely / invalid  & \textipa{t\super Q\super Q} & $330$ & $758$K \\
         \bottomrule
    \end{tabular}
    \caption{Documenting the frequency of phones in \ipapack{}-\textsc{Fleurs}. A phone is considered valid if it is contained in the \texttt{panphon} database \citep{mortensen2016panphon}.}
    \label{tab:phone-breakdown}
\end{table}
\paragraph{Overview.} The dataset comprises  phonetically-transcribed speech for $77$ languages. The recordings and orthographic transcripts were provided by Fleurs \citep{conneau2023fleurs}, while the conversion of the orthographic transcripts to phonetic ones was done by \citet{zhu-etal-2024-taste}.\footnote{The \ipapack{} builds on data from three prior datasets: MSWC \citep{mazumder21_interspeech}, DoReCo \citep{paschen2020building}, and Fleurs \citep{conneau2023fleurs}. We focus on auditing the Fleurs partition since it is the largest in terms of the number of languages.} Each language has at least 3 hours of recordings ($M=10.12$H, $SD=2.74$H). Individual recordings are at most 30 seconds ($M=12.14$s, $SD=1.68s$). We next document and preprocess the contents of the transcripts that are paired with the recordings. 

\paragraph{Transcript composition.}  We segment each phonetic transcript into individual phones, using the \texttt{lingpy} tokenizer \citep{list2016lingpy}.
We count the occurrence of each phone in each language and categorize the phones into the taxonomy in \Table{tab:phone-breakdown}.

We find that the majority of the tokens ($97.0\%$) are valid ones. The majority of these specify a primary place and manner of articulation. 
Moreover, there are a number of phones that have one or two diacritics, for example v\textsuperscript{j}, indicating palatalization. Overall, the dataset contains a high degree of phonetic diversity.

\paragraph{Transcript normalization.} There is however a long tail of $330$ unrecognized phonetic strings (according to the \texttt{panphon} database) out of a total $786$ phone types. This represents 3.0\% of the total tokens in the corpus (approx. $758\text{K}/25\text{M}$). Some of these are invalid unicode representations (e.g., ASCII g is different from the IPA velar plosive \textipa{g}; $31,525$ occurrences). Other times, there are repetitions of diacritics, e.g., \textipa{t\super Q\super Q} ($129$ occurrences). Another common error is non-standard diacritics, e.g., \textipa{o\super N} (instead of \textipa{oN}; 264 occurrences). We correct these invalid phones manually.  After our vocabulary cleaning, we obtain a dataset with $473$ unique, valid phones. We also document the mapping from an invalid phone to a valid phone.\footnote{The mapping is available \href{https://anonymous.4open.science/r/audit-ipa-65BF/replacement_dropbox.json}{here}. We note that there are rare cases of some phonemes such as z\textsuperscript{h} (6 occurrences) that are attested \citep[][]{jacques2011panchronic} (though rare), but considered invalid in \texttt{panphon}. However, such occurrences are rare in both type and token frequency. For example, z\textsuperscript{h} only occurs 6 times.}


The occurrence of invalid or implausible phones in the dataset may be an artifact of the G2P models applied for converting the orthographic transcripts in the Fleurs dataset to the phonetic transcripts in \ipapack{}. For example, one of the Grapheme-to-Phoneme (G2P) models, CharsiuG2P \citep{Zhu2022ByT5MF}, was a byte-level neural model with no constraints mandating that only valid phones be predicted.

\subsection{Preference Proportion Test (PPT)} \label{sec:quality-audit}
We now introduce the statistical test for assessing whether a language subset in \ipapack{} is of reliable quality. Specifically, we ask whether the transcripts are reasonably descriptive of the recordings in the dataset. The phonetic transcripts were automatically generated using G2P models from the orthographic transcripts and thus there is a distinct possibility that the phonetic transcripts do not closely reflect the recording. As seen in prior work on multilingual G2P conversion, performance is far from uniform across language varieties \citep[e.g., the error rate for Egyptian Arabic is more than quadruple that of Spanish in some models;][]{Zhu2022ByT5MF}. 

\paragraph{Problem setup.} Formally, we have a dataset $\mathcal{D}$ with $L$ partitions, one for each language in \ipapack{} -- $\mathcal{D}_1,\dots,\mathcal{D}_{L}$.  The dataset $D$ comprises pairs $(x, G(t))$, where $x$ is an audio recording while $G(t)$ is a phonetic transcript generated by applying a G2P model to an orthographic transcript $t$. However, some of the partitions may be corrupted from systematic G2P conversion errors,  making it unreliable for downstream tasks where a tight correspondence between audio ($x$) and phonetic transcript ($G(t)$) is important. We would like to efficiently identify highly unreliable partitions, annotating only a small sample of $\mathcal{S}_i \subset \mathcal{D}_i$, where $\lvert \mathcal{S}_i \rvert << \lvert \mathcal{D}_i \rvert$. We first describe the setup of the annotation for each datapoint $(x,G(t))$, followed by the construction of $\mathcal{S}_i$. 

\noindent \textbf{Annotating the quality of a transcript}. Directly annotating the quality of a phonetic transcript $G(t)$ in its correspondence with the audio $x$ is challenging, as there is no reference baseline for what makes a transcript high quality. 
Instead, we turn the task to one of pairwise comparison, by asking an annotator to choose between two transcripts: the ground-truth ($G(t)$) or one generated by a reasonably-good quality phone-recognition model ($\mathcal{M}(x)$), like that of \citet{xu2021simple} or \citet{taguchi2023universal}. Eliciting preferences through comparisons rather than absolute judgments has been championed in other work, most notably in Reinforcement Learning from Human Feedback, where others have commended the strategy for providing consistent choices \citep{Christiano2017DeepRL}.
In \Figure{fig:preference}, the transcripts deviate from one another -- for example, the top transcript begins with a rhotic while the bottom begins with a lateral -- and the annotator can listen to the recording to determine which transcript is more faithful.
\\

\paragraph{Hypothesis testing.}  
In order to efficiently determine whether $\mathcal{D}_i$ is an unreliable partition of \ipapack{}, we annotate only a subset $\mathcal{S}_i$. Intuitively, if the annotator consistently prefers the model-generated transcript over the \ipapack{} ground-truth version, ($\mathcal{M}(x) \succ G(t)$), then we may want to discard $\mathcal{D}_i$ from applications in downstream tasks until the transcripts are improved. 

Although $G(t)$ is considered the gold-standard transcript in \ipapack{}, both $G(t)$ and $\mathcal{M}(x)$ are essentially (error-prone) predictions for the phonetic transcript of $x$. We can thus assess the reliability of $G(t)$ by performing a model comparison hypothesis test \citep{card-etal-2020-little} between the two approximations. Specifically, the \textbf{null hypothesis} is that the annotator has no preference for the ground-truth transcript  $G(t)$ over the model-generated one $M(x)$, and the \textbf{alternative hypothesis} is that $G(t)$ is significantly unfavourable. We can model the degree to which the annotator prefers the ground-truth \ipapack{} transcripts as $\theta_G$, where $0\leq \theta_G \leq 1$. When $\theta_G << 0.5$, we can conclude that $\mathcal{D}_i$ is an unreliable dataset for language $i$.\footnote{We choose the null hypothesis to be no preference ($\theta_G=0.5$) because $M(x)$ can for certain languages serves as a very strong baseline. For example, as we show empirically in \Sectref{sec:experiments}, \citet{xu2021simple} and \citet{taguchi2023universal} train very strong models for English and Japanese, respectively. Thus, it is plausible that the annotator would be unbiased with respect to $G(t)$ compared to $M(x)$, even if $G(t)$ was reliable for these two languages.}  We refer to this as the Preference Proportion Test, or the PPT.

\begin{lstlisting}[language=Python, caption={Prints the critical value ($k$) and statistical power of a hypothesis test under a hypothetical effect size (\texttt{theta\_alt} vs. \texttt{theta\_null}) and tolerance for false positives ($\alpha$), for varying numbers of trials ($n$).},label={lst:power}] 
from scipy.stats import binom
def ppt_sample_size(alpha, theta_null, theta_alt):
    for n in range (5, 100, 5): # num. trials
        k = binom(n, theta_null)
            .ppf(alpha) - 1
        power = binom(n, theta_alt)
            .cdf(k)
        print(f"Number of samples: {n};"
            f"Critical value: {k};"
            f"Power: {power}")
\end{lstlisting}

\begin{figure}
    \centering
    \includegraphics[scale=0.5]{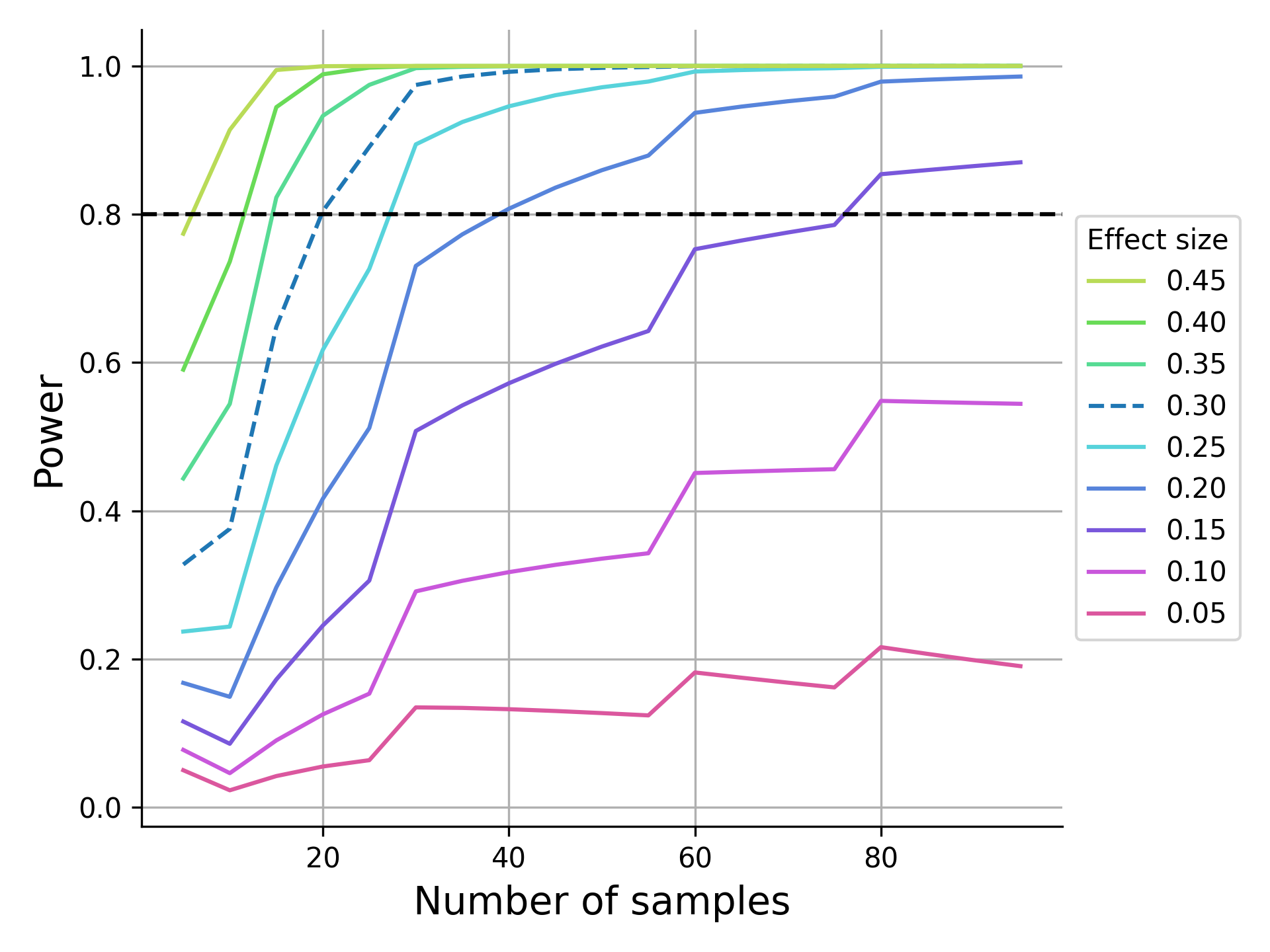}
    \caption{Number of samples that have to be annotated for the Preference Proportion Test (PPT; Sec.~\ref{sec:quality-audit}) for different effect sizes, assuming a fixed false positive tolerance of $\alpha=0.05$. We aim to achieve at least $80\%$ statistical power. We use an effect size of $0.3$ for auditing the \ipapack{} dataset; see Sec.~\ref{sec:ppt-application}.}
    \label{fig:power-analysis}
\end{figure}

For setting the size of the subset to annotate $\mathcal{S}_i$, we perform a power analysis \citep{cohen1992statistical}. We provide the code for the analysis in Listing~\ref{lst:power}. The power analysis is a function of three arguments. First, the tolerance for false positives $\alpha$. Next, the effect size, which is defined by two parameters: the difference between the preference ratio under the null hypothesis ($\theta^{0}_G$) and the alternative hypothesis ($\theta^{A}_G$). We can then supply all three parameters, for example \verb|ppt_sample_size(0.05, 0.5, 0.2)|. We can then test a wide range of potential sample sizes (\verb|for| loop on line 4). For each sample size, we can determine the critical value ($k$; line 5), under which the null hypothesis (equal preference for ground-truth or model-generated) would be rejected. We must also compute the power of this test (line 6), which tells us how likely this outcome is under the alternative hypothesis (model-generated is preferable).

As illustrated in \Figure{fig:power-analysis}, our statistical power increases with the effect size ($\theta^{0}_{G}-\theta^{A}_{G}$) or the number of samples. When we suspect some language subsets to have considerable quality impairments, we can use a large effect size (small  $\theta^{A}_G$) and detect such subsets by annotating only a few samples.

\subsection{Applying the PPT to \ipapack{}} \label{sec:ppt-application}
We now aim to identify any unreliable language partitions $D_i$ of the \ipapack{} dataset with the Preference Proportion Test (PPT). We describe the selection of the language subsets to audit, the number of samples to annotate in each subset, and how we format the datapoints to elicit annotations.

\begin{figure}
    \centering
    \includegraphics[scale=0.5]{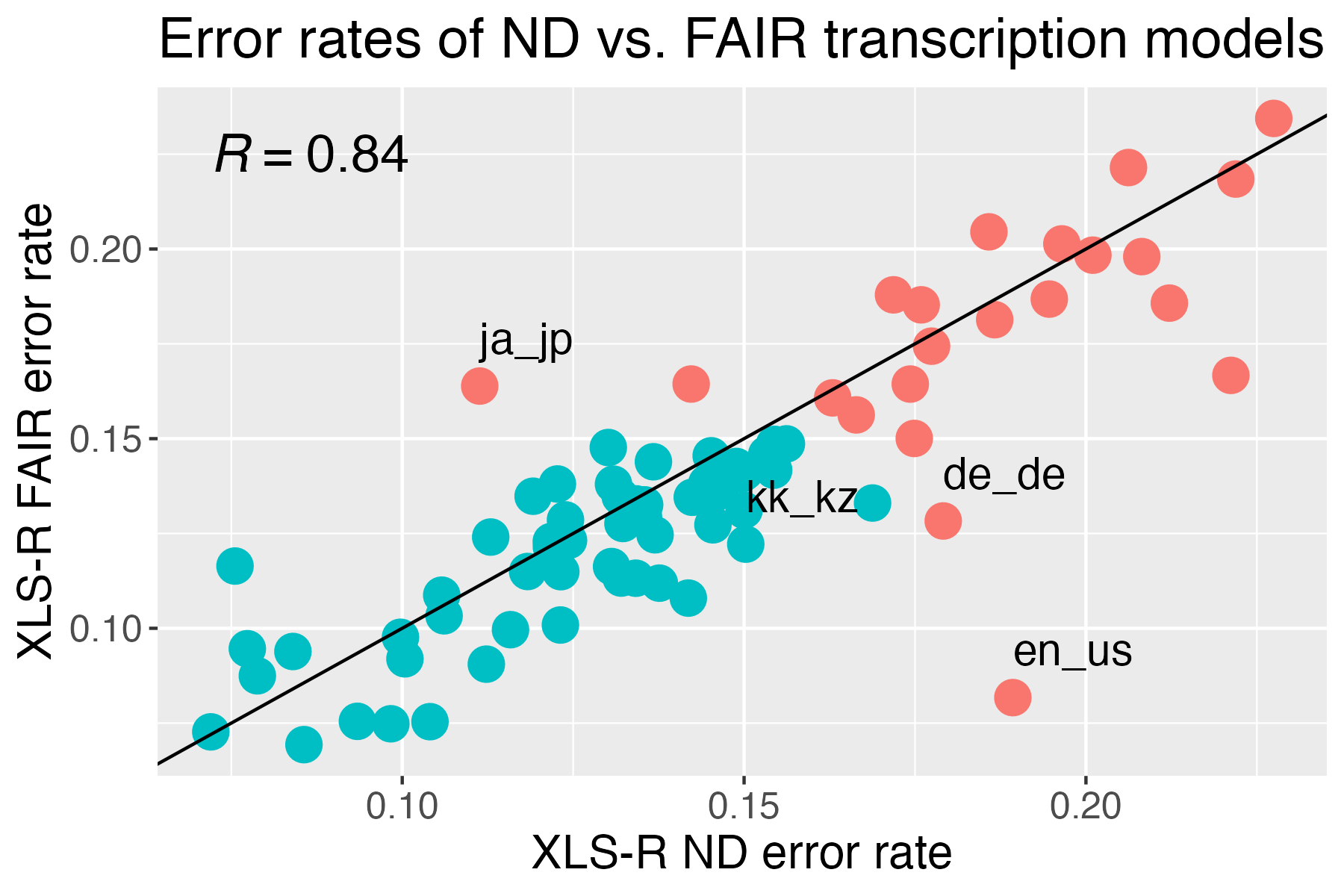}
    \caption{Benchmarking the models of \citet{taguchi2023universal} (X-axis) and \citet{xu2021simple} (Y-axis) on \ipapack{}. Error rates  for both models are measured by phonetic feature discrepancies \citep{mortensen2016panphon} between the model-generated transcripts and the transcripts in \ipapack{}. In \Sectref{sec:quality-audit}, we audit the languages with relatively high-error rates (top-right quadrant, in red), to determine whether the errors may be attributed to poor-quality transcripts in \ipapack{}.} 
    \label{fig:error-rate}
\end{figure}
\paragraph{Selecting languages to audit.} 
We first select a subset of languages in X-IPAPack to audit.
To do so, we apply leverage existing phone recognizers, and select languages where the recognizer predictions have a high rate of discrepancies compared to the gold-standard transcripts. 
To compute the discrepancy, 
we follow \citet{taguchi2023universal} and
use the Phonetic Feature Error Rate (PFER) with phonetic feature vectors from \texttt{panphon} \citep{mortensen2016panphon}. We normalize the error rates by the length of the \ipapack{} transcript in terms of number of phones, ensuring that the variation in error rates is not an artifact of length. In \Figure{fig:error-rate}, we plot the error rates for all languages in \ipapack{} using the phone recognizers of \citet{taguchi2023universal} and \citet{xu2021simple}, which we denote \nd{} and \fair{} respectively. We use these recognizers as they demonstrated competitive performance against human transcribers in recent work \citep{taguchi2023universal}.

We find that there is a substantial amount of variation in performance across languages for both models. In the top right of \Figure{fig:error-rate}, language datapoints coded in red (e.g., Malayalam; length-normalized PFER: $0.20$) achieve error-rates more than double that of languages on the bottom left (e.g., Swahili; length-normalized PFER $0.08$). As shown in \Figure{fig:error-rate}, this variation is mostly robust to the choice of the recognition model, which have a correlation of $r=.84$. We thus select all languages with an error rate in the third quantile of either model ($.15$ for \citet{xu2021simple} and $.17$ for \citet{taguchi2023universal}) for annotation with the PPT, $L=22$ languages in total. For each language, we generate transcripts using the model that performs better for that language, for example \citeauthor{xu2021simple} for English and \citeauthor{taguchi2023universal} for Japanese.

\paragraph{Setting parameters for PPT.} We are auditing for $D_i$ where there are systematic discrepancies between the audio $x$ and the ground-truth transcript $G(t)$, so the value of $\theta^{A}_{G}$ for the alternative hypothesis should be much lower than the null hypothesis $\theta^{0}_{G}$ value of $0.5$.  We thus set $\theta_G^{A}=0.2$ for the alternative hypothesis, an effect size of $\theta^{0}_{G} - \theta_G^{A}=  0.3$.\footnote{Since this is a one-sided hypothesis test, our test will detect $\theta_G\leq 0.2$.}  With a false positive error tolerance of $\alpha=0.05$, we find that we can achieve a test with a statistical power of $80.4\%$ through annotating $n=20$ samples (\Figure{fig:power-analysis}), which is considered a high-powered test \citep{card-etal-2020-little}.
Specifically, we annotate the samples and reject the null hypothesis when only $k=5$ times or fewer do we prefer the \ipapack{} transcript $G(t)$ over the model-generated one $M(x)$.\footnote{It would also be plausible to select a slightly smaller or slightly larger $\theta_G$ for the alternative hypothesis -- this would only require slightly fewer or more samples to be annotated, respectively. The overall point is that for detecting large quality dropoffs for $G(t)$ (relative to $M(x)$), only $\lvert S_i\rvert$ rather than $\lvert D_i \rvert$  samples need to be annotated.}   

\noindent \textbf{Aligning gold-standard and baseline transcripts}. Unlike an \ipapack{} transcript $G(t)$ that contains space-delimited phone strings, the corresponding phonetic language model transcript $\mathcal{M}(x)$ is a phone string with no spaces. In order to facilitate the comparison of the two transcripts, we induce spaces in the model-generated transcript using the Needleman-Wunsch alignment algorithm \citep{needleman1970general,kleinberg2006algorithm} to align the model transcript with the transcript.\footnote{We could also remove spaces from both transcripts, but this makes the transcripts very difficult to read.} Since the algorithm requires computing the similarity between pairs of phones, we employ articulatory feature vectors from \texttt{panphon}. That is, we encode phones as binary feature vectors -- for example, whether the phone is voiced or unvoiced) -- enabling a graded measure of similarity by computing the hamming distance between the vectors \citep{mortensen2016panphon,taguchi2023universal}.

\paragraph{Annotation process.} 
For every datapoint, the annotator selects between $G(t)$ and $M(x)$ as the preferable transcript for an audio recording from \ipapack{}. We demonstrate the annotation interface and instructions in \Appendix{app:annotation-interface}.
They could replay the recording as many times as desired before making their choice, and could configure the recording to play at normal speed or at $0.25/0.50/0.75$ speed. 
Both $G(t)$ and $M(x)$ reflected a broad or phonemic transcription style, as they are either the result of G2P conversion (in the case of $G(t)$) or were trained on G2P transcripts (in the case of $M(x)$). The annotator completed all $20$ samples for a language partition $S_i$ from \ipapack{} before moving onto the next data subset $S_j$ ($j\neq i$).


\begin{figure}
    \centering
    \includegraphics[scale=0.50]{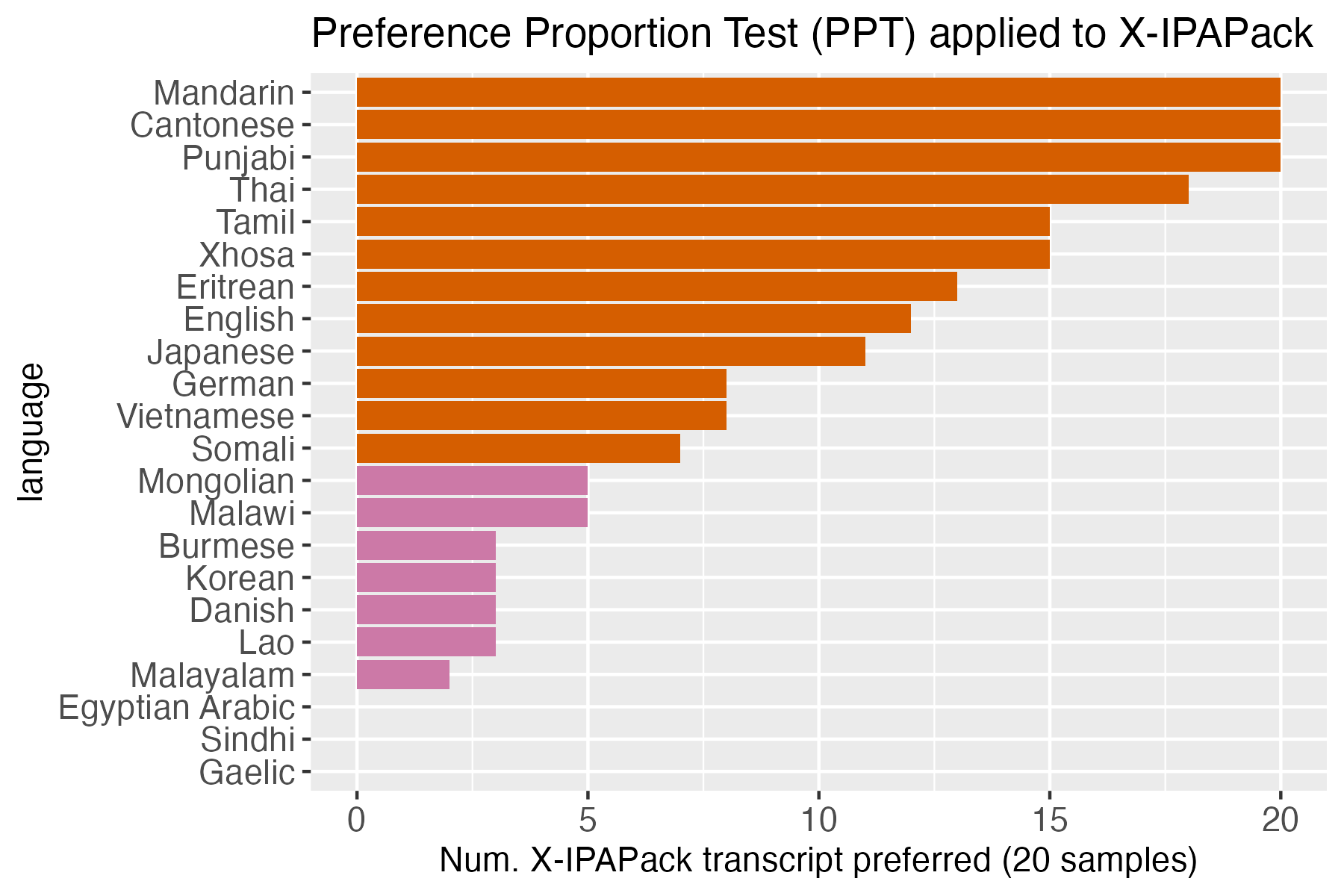}
    \caption{Number of samples where the \ipapack{} transcript was preferred to one generated by a phone recognizer. The bottom $10$ languages have a relatively large number of samples where the \ipapack{} transcript is of lesser quality than the phone recognizer transcript, and thus they fail the Preference Proportion Test.}
    \label{fig:ppt-proportions}
\end{figure}

\begin{table}[]
    \centering
    \begin{tabular}{l|lll}
        \textbf{Language} & $G(t)$ & $\mathcal{M}(x)$ \\
        \toprule 
         Egyptian Arabic & \textipa{taStahir} & \textipa{teSteher} \\
         Malayalam & \textipa{\textteshlig a:rinte} & \textipa{\textteshlig a:Ri\textrtailn\textrtaild i} \\
         English & \textipa{tuflaI} & \textipa{t@flaiI} \\
         \bottomrule
    \end{tabular}
    \caption{Examples of cases where the model-generated transcript $\mathcal{M}(x)$ is preferable to the ground-truth transcript $G(t)$ from \ipapack{}. Note that transcripts and predictions are generally longer (full sentences); see \Figure{fig:preference} for a realistic example of the annotation task.}
    \label{tab:pref-examples}
\end{table}
\paragraph{Identifying unreliable language partitions.} \Figure{fig:ppt-proportions} shows that out of the $L=22$ languages, we reject the null hypothesis for $10$, indicating that the \ipapack{} transcript for these languages is unreliable. We demonstrate examples where we prefer the model in \Table{tab:pref-examples}. For example, in Egyptian Arabic ($\theta_G=0/20$), we find that the vowels are systematically mis-transcribed, often substituting open vowels (\textipa{a}) for close-mid ones (\textipa{e}). In Malayalam ($\theta_G =2/20$), we find that the transcripts regularly mistake retroflex consonants for alveolar ones, in addition to misidentifying voicing (\textipa{t} instead of \textipa{d}). We also find that the model-generated transcript $\mathcal{M}(x)$ may be preferable over the ground-truth transcript even when the ground-truth transcript $G(t)$ reasonably represents the audio recording. For example, for American English ($\theta_G = 12/20$), while the \ipapack{} transcripts are generally reliable, $\mathcal{M}(x)$ often provides a more faithful transcription since it can identify cases of co-articulation or reduction (\textipa{@} instead of \textipa{e}).


\paragraph{Agreement on the PPT annotations.} In order to assess the agreement that the flagged languages were indeed of low quality, we had a second co-author independently annotate $50$ datapoints. Specifically, they annotate $10$ datapoints for $5$ of the flagged languages: Egyptian Arabic,  Danish, Gaelic, Malayalam, Sindhi. We find that the annotators largely agree, providing the same preference on $43/50$ datapoints.

\paragraph{Summary.}  We demonstrate that annotation of $20$ samples per language can enable identifying unreliable language partitions in multilingual datasets. Using our PPT procedure, we efficiently identify a number of languages with low quality transcripts (Malayalam, Egyptian Arabic, among others) in the \ipapack{} dataset. We thus recommend that these partitions be omitted from use until their phonetic transcripts are remediated.  


\section{Data Quality Effect on Downstream Performance} \label{sec:experiments}
We now assess the downstream effect of removing the $L'=10$ unreliable language partitions $\mathcal{D}_1,\dots, \mathcal{D}_{L'}$ from the \ipapack{} dataset $\mathcal{D}$. We do so by training two phone recognition models, one on $\mathcal{D}$ and one on $\mathcal{D}-(\mathcal{D}_1\cup\dots\cup \mathcal{D}_{L'})$. More specifically, we finetune two Whisper (small) models on these datasets. We refer to the former as \whisperipa{} and the latter as \whisperipadist{}. We provide hyperparameter details in \Appendix{app:hyperparameters}. To contextualize our results, we also compute performance for two other models that demonstrated strong performance on multilingual phone recognition: \citet[\nd{}; ][]{taguchi2023universal} and \citet[\fair{}; ][]{xu2021simple}.

For our evaluation dataset, we select $5$ languages from \Sectref{sec:ppt-application} that passed the PPT, indicating they are trustworthy for evaluation. Specifically, we evaluate on a test set \dtest{} comprising the \ipapack{} evaluation partitions for Thai, English, Japanese, Punjabi, and Eritrean. We hold these languages out from training of the two Whisper-based phone recognition models. To evaluate our models, we again compute the Phonetic Feature Error Rate (PFER) using phonetic feature vectors from \texttt{panphon} \citep{mortensen2016panphon}. 

\begin{table}[]
    \centering
    \begin{tabular}{p{1.2cm}|p{0.45cm}p{0.45cm}p{0.45cm}p{0.45cm}p{0.55cm}|p{0.40cm}}
         \textbf{Model} & \textbf{Th.} & \textbf{En.} & \textbf{Ja} & \textbf{Pa.} & \textbf{Ti.} & \textbf{Avg} \\
         \toprule 
         \fair{} & 20.3 & \textbf{7.0} & 15.9 & 15.1 & 18.2 & 15.3 \\ 
\nd{} &  19.0 & 15.2 & \textbf{10.5} & 15.2 & 18.6 & 15.7 \\ 
\whisperipa{} & \textbf{15.1} & 13.7 & 11.5 & 13.8 & 16.6 & 14.1 \\ 
\whisperipadist{} & 15.2 & 12.3 & 11.2 & \textbf{11.0} & \textbf{16.4} & \textbf{13.2} \\ 
         \bottomrule
    \end{tabular}
    \caption{Performance on held-out languages from IPAPack-Fleurs, as measured by Phonetic Feature Error Rate (Median). We use the \textit{small} variant of the Whisper model. The macro-average is computed by averaging the error rate across all 5 languages, with each language given equal weight.}
    \label{tab:fleurs-held-out}
\end{table}

\subsection{Results} \label{sec:experiment-results}
In \Table{tab:fleurs-held-out}, we demonstrate the performance of the four models -- \fair{}, \nd{}, \whisperipa{}, \whisperipadist{} -- on the $5$ held out languages. Averaging across the $5$ datasets, we find that \whisperipadist{} achieves the best performance. Importantly, it improves upon \whisperipa, despite the the latter being trained with more datapoints (drawn from the low-quality transcript languages identified in  \Sectref{sec:quality-audit}) and $12\%$  more optimization steps ($3982$ vs. $4466$, respectively, for completing $2$ epochs of training on their respective datasets). 

\paragraph{Low-quality data impairs performance.} Comparing \whisperipa{} and \whisperipadist, we find the largest improvement on Punjabi ($13.82$ vs. $11.01$ PFER), a relative improvement of $20.3\%$. One likely reason for this sizeable improvement is that through the PPT (\Sectref{sec:quality-audit}), we pinpointed in \ipapack{} a related language -- Sindhi -- that contained low-quality transcripts, and removed it from the training set for \whisperipadist{}. Both Punjabi and Sindhi are Indo-European languages \citep{wals}. Malayalam may have also had an impact. Although it is from a different language familiy (Dravidian), it is also spoken in the Indian subcontinent and may share some common phonetic features from broad areal effects \citep{everett2015climate}. We also observe improvements (albeit smaller ones) on English, Japanese, and Eritrean, with Thai being the only language where \whisperipa{} achieves a slightly higher PFER over \whisperipadist{}. 

\paragraph{Multilingual phone recognition models are not language-agnostic.} Comparing \whisperipadist{} with \fair{} and \nd{}, we find that the former achieves better performance in all but two: English for \fair{} and \nd{} for Japanese. This is likely due to \fair{} having English in its training set \citep{xu2021simple} and \nd{} having Japanese in its training set \citep{taguchi2023universal}.\footnote{It's worth noting that Japanese is also in the \fair{} training set. Despite this, its performance compared to \whisperipadist{} and \nd{} is considerably worse.} Moreover, when we train another \whisperipadist{} on all languages, including \ipapack{}-English, we obtain better performance on English than \fair{} (PFER of $6.95$ vs. $6.32$). Thus, we find that despite all the models having been trained on a reasonably diverse set of languages, performance still varies depending on the exact training data composition.

\subsection{Evaluation on \textsc{IPAPack-DoReCo}} \label{subsec:eval-doreco}
Next, we evaluate the models on another partition of \ipapack{}, the DoReCo partition  \citep[rather than the \textsc{Fleurs} partition we have been using up to this point,][]{zhu-etal-2024-taste} to further assess model capabilities on an out-of-distribution set of languages. \ipapack{}-DoReCo comprises phonetically transcribed speech for $44$ endangered languages \citep{paschen2020building}. The utterances tend to be shorter than \ipapack{}-Fleurs \citep{zhu-etal-2024-taste}. Note that we don't apply the PPT on \ipapack{}-DoReCo, since its construction had oversight from expert linguists for each language \citep{zhu-etal-2024-taste,paschen2020building}. This evaluation sheds further light on model performance since these languages were never seen during finetuning for any of the models; they are also highly unlikely to have been observed during the multilingual pretraining stages for Whisper \citep{radford2023robust} and XLS-R \citep{babu2021xls}.  

\begin{table}[]
    \begin{tabular}{p{1.5cm}p{1.55cm}p{1.0cm}r}
         \textbf{Model} & \textbf{PFER (Median)} & \textbf{Var. (IQR)} & \textbf{Best} \\
         \toprule
         \fair & $\mathbf{5.19}$ & $4.21$ & $\mathbf{29/44}$ \\
         \nd & $5.48$ & $4.18$ & $5/44$\\
         \whisperipadist{} & $5.60$ & $4.39$ & $10/44$\\
         \whisperipa{} & 7.54 & $66.89$ & 0/44\\
         Allosaurus & $7.53$ & $5.04$ & $0/44$\\
         \bottomrule
    \end{tabular}
    \caption{Evaluation results on \ipapack{} (44 languages). PFER: Phonetic Feature Error Rate; \citep{mortensen2016panphon}. IQR: Interquartile Range. Highest refers to the number of times the model performed the best within one of the \ipapack{}-DoReCo languages.}
    \label{tab:doreco-results}
\end{table}
\paragraph{Poor data quality impairs out-of-distribution performance.} In \Table{tab:doreco-results}, we see that \whisperipadist{} 
significantly outperforms \whisperipa{} (PFER $5.60$ vs. $7.54$, representing a $25.7\%$ error rate improvement). We emphasize that the improvement arises solely from having removed the low-quality language subsets from finetuning  (\Sectref{sec:ppt-application}), as no other factors were manipulated. We find that trained on the unfiltered \ipapack{} dataset, \whisperipa{} performs no better than a much older model, Allosaurus. We also find that \whisperipa{} is prone to entirely degenerate predictions, such as empty strings and predictions of the same character ad-nauseam, leading to some predictions that incur an extremely high error and a high variance (Interquartile Range; IQR) of $66.89$. \whisperipadist{} is also susceptible to degenerate predictions, but to a much lesser degree, given its reasonable IQR ($4.39$). Overall, our results suggest that degenerate predictions are exacerbated by low-quality training data.\footnote{All of our experiments use greedy generation for all models. We did not find substantive improvements from beam search.}

\paragraph{\fair{} achieves strongest out-of-distribution performance, though there is language-conditional variance.} In \Table{tab:doreco-results}, taking the median across all datapoints in all $44$ languages, we find that \fair{} achieves the best performance, with a median PFER of $5.19$. \whisperipadist{} and \nd\ achieve similar performance (PFER $5.60$ and $5.48$, respectively). 

Overall, while \fair{} exhibits dominance in this test split, we note that there is language conditional variability. There are $10/44$ and $5/44$ languages where \whisperipadist{} or \nd{} obtain better performance. 

\section{Dataset Coverage Limits Downstream Performance} \label{subsec:phone-level-error}
Our analyses in training phone recognizers in \Sectref{sec:ipapack} demonstrated that poor-quality data impairs generalization. However, filtering out low-quality data from multilingual datasets does not necessarily guarantee multilingual generalization. Prior work had claimed universal phone recognition capacity from training on a small number of languages on high-quality data \citep{taguchi2023universal}, but we demonstrate that multilingual generalization remains limited by the training data composition. The number of attested sounds across the world's languages is large \citep{moran2019phoible}, and their frequencies have a Zipfian distribution \citep{macklin2020re}, making them challenging to learn from limited data. We demonstrate these challenges through two phone segment-level error analyses.

\begin{table}[]
    \centering
    \begin{tabular}{cp{0.75cm}|p{1.5cm}p{1cm}p{1cm}}
    \textbf{Click} & \textbf{Freq.} & \textbf{\whisperipadist{}} & \textbf{\fair} & \textbf{\nd} \\
    \toprule
    \textipa{||} & $35$ & $\mathbf{.83}$ & $0.00$ & $0.00$\\
    \textipa{|} & $87$ & $\mathbf{.73}$ & $0.00$ & $0.00$\\
    \textipa{!} & $102$ & $\mathbf{.89}$ & $0.00$ & $0.00$\\
    \bottomrule
    \end{tabular}
    \caption{Recall on predicting click consonants. Only the model trained on \ipapack{} is able to predict clicks, which manifest in Zulu and Xhosa in the \ipapack{} data. }
    \label{tab:click-predictions}
\end{table}

\paragraph{Purported universal phone recognizers cannot transcribe clicks.} Since \fair{} and \nd{} were not trained on any Bantu languages or any other language or dialect containing click consonants, we find they are incapable of predicting clicks in \Table{tab:click-predictions}. By comparison, \ipapack{} contains Zulu and Xhosa transcripts. Moreover, they pass the PPT test (\Sectref{sec:quality-audit}), indicating that they are reliable. Indeed, we find that \whisperipadist{} performs well at identifying clicks in the evaluation dataset for these languages when they are present in the recording. This demonstrates training a universal phone transcription model is more challenging than previously thought, since it requires accurate identification of typologically rare sounds that may not be prevalent in prior datasets.

To provide finer-grained measurements of efficacy at recognizing a certain sound, we use the \textbf{Expected Phone Error (EPR)} metric, defined as follows. Given a phone $q$, we identify all its occurrences in the ground-truth transcripts $G(t)$. Assuming that $q$ appears at position $i$ in transcript $G(t)$, we then compute the error as the phonetic feature distance \citep[using \texttt{panphon};][]{mortensen2016panphon} between $G(t)_i$ and the phone in $\mathcal{M}(x)$ that is aligned to $G(t)_i$. We then average the error from each occurrence of $q$ in the dataset. We apply the Needleman-Wunsch algorithm for the alignment.



\begin{table}[]
    \centering
    \begin{tabular}{p{1cm}p{0.7cm}p{3cm}p{1cm}}
         Tgt. phone &  Lang. & EPR (\whisperipadist{}/FAIR/ND) & Maj. label \\
         \toprule
         \textrtailt & pa\_in & $\mathbf{0.09}/0.25/0.21$ & \textrtailt, t,  \textrtailt\\
         \textipa{\textrtailt \super h} & pa\_in & $\mathbf{0.14}/0.21/0.36$ & \textrtailt\super h, t, k\\
         \textrtaild & pa\_in & $\mathbf{0.15}/0.26/0.19$ & \textrtaild, d, \textrtaild \\
         \midrule
         \textipa{D} & en\_us & $0.25 / \mathbf{0.08} / 0.32$ & \textipa{d}, \textipa{D}, \textipa{d} \\
         \textipa{\*r} & en\_us & $0.23 / \mathbf{0.14} / 0.36$ & r, \textipa{\*r}, - \\
         \textipa{I} & en\_us & $0.19 / \mathbf{0.11} / 0.25$ & \textipa{I} , \textipa{I}, i\\
         \midrule
         \textipa{C} & ja\_jp & $0.25 / 0.20 / \mathbf{0.16}$ & \textipa{S} , \textipa{S} , \textipa{C}\\ 
         \textipa{e:} & ja\_jp & $\mathbf{0.04} / 0.13 / 0.10$ & \textipa{e:, e, e:}\\
         \textipa{e} & ja\_jp & $0.07/0.14/\mathbf{0.06}$ & \textipa{e, e, e} \\
         \bottomrule
    \end{tabular}
    \caption{Models vary in their ability to predict certain phones, with \whisperipadist{} better at Punjabi; \nd{} at Japanese; and \fair{} at English. EPR: Expected Phone Error for the three models \textbf{(\whisperipadist{}/\fair{}/\nd{})}. Maj. label: the phone most commonly predicted by each of the three models (same order as EPR).}
    \label{tab:phone-level-errors}
\end{table}

\paragraph{Models vary widely in the phonetic details they can capture.} In \Table{tab:phone-level-errors}, we show three languages where one of the three models excels. For Punjabi, we see that \whisperipadist{} is highly effective at identifying retroflex consonants, even distinguishing between aspirated and unaspirated stops (\textipa{\textrtailt \super h} and \textrtailt), an important phonemic distinction in Punjabi \citep{jain2007indo}.
By comparison, \fair{} predicts unmarked alveolar stop (\textipa{t}) in both cases.  For Japanese, we find that both \whisperipadist{} and \nd{} can distinguish between \textipa{e} and \textipa{e:} -- another phonemic distinction -- while \fair{} cannot. For English, we find that \fair{} is significantly better at distinguishing between dental stops and fricatives, as well as high-front and near-high-front vowels (\textipa{i} and \textipa{I}) than the other two models. This is unsurprising since it is the only model with English in its training dataset. 

\paragraph{Summary.} Our results indicate that the training data composition remains relevant; training models even on fairly large multilingual datasets \citep{xu2021simple} does not enable fine-grained cross-linguistic generalization. Thus, acquiring language-specific training data remains important. Moreover, our results in \Sectref{sec:experiment-results} demonstrate that a systematic audit of the quality of acquired data is highly beneficial for downstream performance. Our Preference Proportion Test (\Sectref{sec:quality-audit}) enables systematic and sample-efficient quality auditing for this purpose. 


\section{Conclusion}
We present the Preference Proportion Test (PPT; \Sectref{sec:quality-audit}) for efficiently and systematically auditing the quality of data from specific languages in a large multilingual dataset. We apply the PPT for efficiently identifying low quality language partitions in the recent \ipapack{} dataset \citep{zhu-etal-2024-taste}. Our audit is effective, efficiently identifying language subsets whose complete removal brings substantial improvements in the downstream task of automatic phone recognition. Auditing the quality of multilingual datasets is critical for stewardship of the multilingual datasets in the field, ensuring that they are reliable, trustworthy, and representative \citep{kreutzer-etal-2022-quality}. Overall, our method lays the foundation for statistically principled multilingual dataset auditing.

\bibliography{tacl2021}
\bibliographystyle{acl_natbib}


\newpage
\appendix
\section{Hyperparameter Settings}
\label{app:hyperparameters}
\begin{table}[]
    \centering
    \begin{tabular}{l|l}
    \textbf{Hyperparam}. & \textbf{Setting}\\
    \toprule
         Batch size & $64$  \\
         Num. epochs & $2$\\ 
         Learning rate & $1e-4$\\
         FP16 & True\\
         Max gen. length & $225$\\
         Grad. checkpointing & True\\
         Warmup steps & $500$\\
         \bottomrule
    \end{tabular}
    \caption{Hyperparameter settings for finetuning the Whisper models in \Sectref{sec:experiments}.}
    \label{tab:hyperparams}
\end{table}
See \Table{tab:hyperparams}. We performed a hyperparameter sweep using the validation sets in \ipapack{} for the evaluation languages. Our hyperparameter sweep was over batch sizes of $\{8, 16, 32, 64, 128, 256\}$, precision of either FP16 or BF16, and warmup steps of $\{500, 1000\}$ and uniformly distributed learning rates of $\{1e-5, 3e-4\}$.

\section{Annotation Interface}\label{app:annotation-interface}

\begin{figure*}
    \centering
    \includegraphics[scale=0.45]{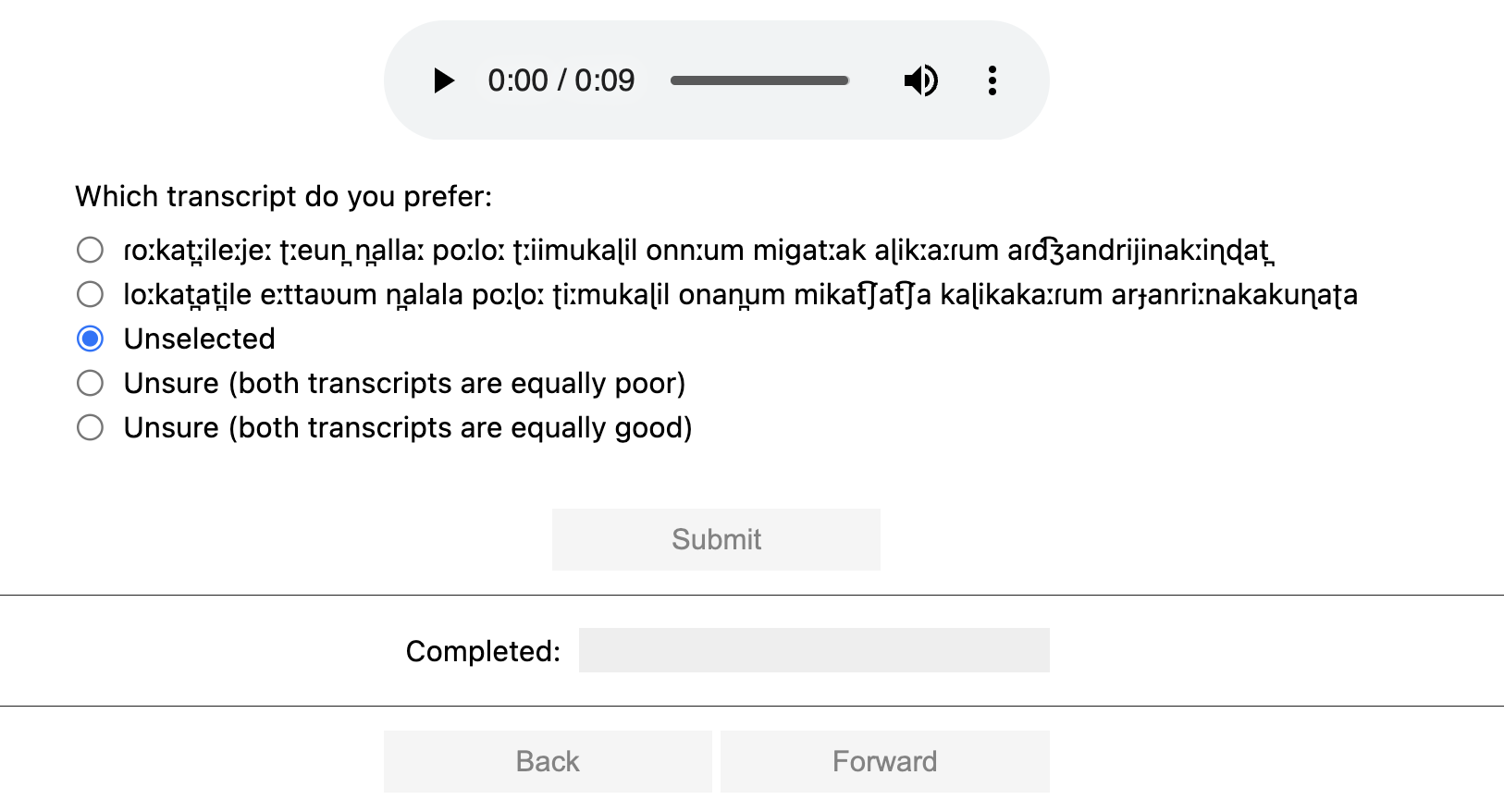}
    \caption{Interface for performing annotations.}
    \label{fig:interface}
\end{figure*}
We show the interface for performing the annotations in \Figure{fig:interface}. The annotator must select from one of four options. One of the transcripts is the gold-standard transcript from \ipapack{}, while the other is generated from a baseline model (either \fair{} or \nd{}). We randomize whether the gold-standard or the baseline prediction is displayed first.

\begin{figure*}
    \centering
    \includegraphics[scale=0.45]{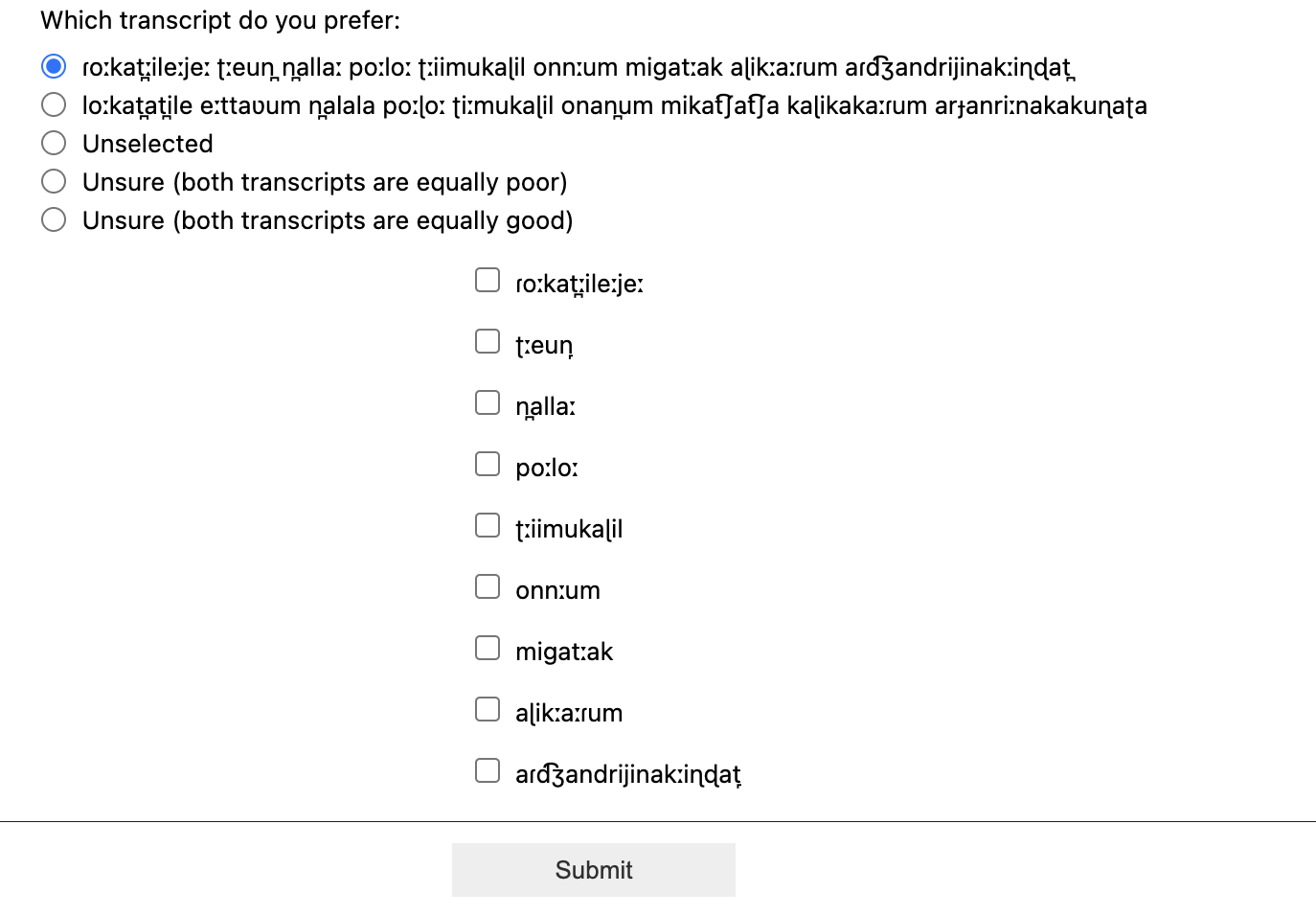}
    \caption{Selecting substrings that were influential in the decision once a decision is made.}
    \label{fig:interface-expanded}
\end{figure*}

As long as the annotator picks either the gold-standard or the baseline model prediction, and not one of the abstention options, they are given the opportunity to select which word(s) most influenced their selection (\Figure{fig:interface-expanded}). We store these selections purely for documentation of the annotation, it does not influence any of the analyses in the main text.

\onecolumn
\begin{tcolorbox}[width=\linewidth]
First, we thank you for agreeing to helping us with providing your preference annotations! Your expertise in phonetic transcription is valuable and appreciated. Please read the following instructions carefully.\\

\paragraph{Options.} The annotation task will comprise 50 datapoints of the form: (audio, transcript1, transcript2). You'll listen to the audio and then pick one of four options:

\begin{enumerate}
\item  TranscriptA > TranscriptB 
\item  TranscriptA < TranscriptB 
\item TranscriptA = TranscriptB (both are equally good)
\item TranscriptA = TranscriptB (both are equally poor)
\end{enumerate}

You may replay the audio as many times as you like. You can also change the playback speed of the audio using the "kebab menu" (three dots).

\paragraph{Justification.} When you select (1), you will then select which word(s) in the transcript were better represented in TranscriptA compared to TranscriptB. (Analogous for Option 2). No need to be exhaustive here, just select some of the word(s) that seemed most well represented to you (relative to the other transcript).

\paragraph{Avoid abstaining (options 3 and 4) if possible.} Try to select option (1) or (2) when possible, only resorting to (3) or (4) when you find it impossible to pick between the two. (Ideally, no more than 10 samples should have the (3) or (4) option). When both TranscriptA and TranscriptB have problems, try to select the transcript that has fewer problems.

\paragraph{Transcript spacing.} When comparing the two transcripts, don't use the whitespace segmentation of the transcripts in informing your decision. The spaces are automatically inserted to improve readability of the transcripts, and there may occasionally be some spacing errors in TranscriptA relative to TranscriptB (or vice-versa). For example, one of the transcripts may segment the phrase "the car" into ``\textipa{ð@k Ar}'' instead of ``\textipa{ð@ kAr}''. Please try to ignore these spacing discrepancies in making your determination, instead focusing on whether the phonetic segments accurately represent the speech audio. For example, if TranscriptA = ``\textipa{Ti kEr}'' while TranscriptB = ``\textipa{D\textipa{@}k Ar}'', you should prefer TranscriptB since it has more accurate phonetic segments (assuming a Standard American English pronunciation), and ignore the fact that the space is inserted after the ``k'' in \textipa{kAr} rather than after the vowel in ``\textipa{D@}''. Note that the spacing can also result in affricates (``\textipa{\textdyoghlig}'') being broken up (``\textipa{d Z}''), so if you clearly hear an affricate but don't see a tie bar in the transcript, the affricate may well be represented but (inadvertently) broken up between two words.

\paragraph{Going back to previous annotations.} The interface contains back and forward buttons. When you hit the back button to navigate to the previous sample, you'll see that your prior annotation is stored. If you want to select a different option (1-4), you can do so and hit submit. If you want to leave it as is, you can just hit the "Forward" button again. When you use "Back", The "Submit" button will be disabled unless you change your selection. 

\paragraph{Display.} On rare occasions, some of the transcripts can be a bit long and rendered incorrectly. Ensuring that the window is full-screen and using a larger external monitor (if possible) should mitigate this.

When you are ready, execute the next cell to begin. At the bottom of the cell, the interface for performing the annotations will appear. Your progress will be saved as you work through the annotations, so feel free to take breaks. 

\end{tcolorbox}

\end{document}